\documentclass{article}

\usepackage{microtype}
\usepackage{graphicx}
\usepackage{subfigure}
\usepackage{booktabs} 
\usepackage{enumitem}

\usepackage{todonotes}

\usepackage{hyperref}

\usepackage[accepted]{preprint_icml2018}

\usepackage{ulem}

\icmltitlerunning{One-Shot Segmentation in Clutter}

\begin{document}

\twocolumn[
\icmltitle{One-Shot Segmentation in Clutter}



\icmlsetsymbol{equal}{*}

\begin{icmlauthorlist}
\icmlauthor{Claudio Michaelis}{tue,bccn}
\icmlauthor{Matthias Bethge}{tue,bccn,mpi,cnai}
\icmlauthor{Alexander S. Ecker}{tue,bccn,cnai}
\end{icmlauthorlist}

\icmlaffiliation{tue}{Centre for Integrative Neuroscience and Institute for Theoretical Physics, University of T\"ubingen, Germany}
\icmlaffiliation{bccn}{Bernstein Centre for Computational Neuroscience, T\"ubingen, Germany}
\icmlaffiliation{mpi}{Max Planck Institute for Biological Cybernetics, T\"ubingen, Germany}
\icmlaffiliation{cnai}{Center for Neuroscience and Artificial Intelligence, Baylor College of Medicine, Houston, TX, USA}

\icmlcorrespondingauthor{Alexander Ecker}{alexander.ecker@uni-tuebingen.de}

\icmlkeywords{Machine Learning, ICML}

\vskip 0.3in
]



\printAffiliationsAndNotice{}  

\begin{abstract}
We tackle the problem of one-shot segmentation: finding and segmenting a previously unseen object in a cluttered scene based on a single instruction example. We propose a novel dataset, which we call {\it cluttered Omniglot}. Using a baseline architecture combining a Siamese embedding for detection with a U-net for segmentation we show that increasing levels of clutter make the task progressively harder. Using oracle models with access to various amounts of ground-truth information, we evaluate different aspects of the problem and show that in this kind of visual search task, detection and segmentation are two intertwined problems, the solution to each of which helps solving the other. We therefore introduce {\it MaskNet}, an improved model that attends to multiple candidate locations, generates segmentation proposals to mask out background clutter and selects among the segmented objects. Our findings suggest that such image recognition models based on an iterative refinement of object detection and foreground segmentation may provide a way to deal with highly cluttered scenes.
\end{abstract}

\section{Introduction}
\label{introduction}

\begin{figure}[t]
\begin{center}
\centerline{\includegraphics[width=\columnwidth]{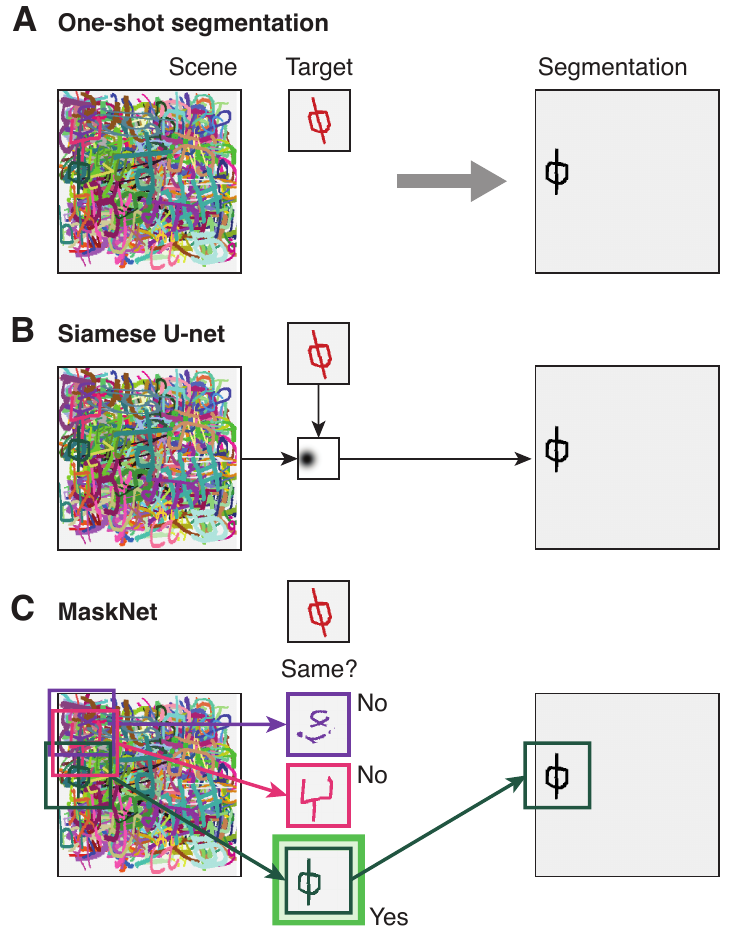}}
\caption{One-Shot Segmentation.
\textbf{A,} Goal: find a {\it target} in a cluttered {\it scene} and produce a pixel-wise segmentation. 
\textbf{B,} Our {\it Siamese U-net} baseline localizes the target, then segments it.
\textbf{C,}~{\it MaskNet} generates proposals of segmented instances, masks the background, then computes the best match.
}
\label{fig:teaser}
\end{center}
\vskip -0.2in
\end{figure}

Humans are not only good at learning to recognize novel, unknown objects from a single instruction example (one-shot learning), but can also localize these objects in highly cluttered scenes and segment them from the background. 

In the computer vision community, one-shot learning has recently received a lot of attention and substantial progress has been made in the context of image classification \cite{Koch2015a, Lake2015, Vinyals2016, Bertinetto2016, Snell2017,Triantafillou2017a, Shyam2017}.
Segmentation, however, is still very much tied to classification, limiting its applicability to datasets with less than a few hundred semantic or object classes (or subsets thereof, e.\,g. the SceneParse150
benchmark on ADE20k \cite{Zhou2017b}).
This stands in contrast to humans who can segment previously unseen objects simply by using contextual information.

In the present paper, we work towards closing this gap by tackling the problem of one-shot segmentation: 
Given a single instruction example (the {\it target}) and a cluttered image with many objects (the {\it scene}), find the target in the scene and produce a pixel-wise segmentation (Fig~\ref{fig:teaser}A).
This task is harder than the multi-way discrimination task often employed for one-shot learning because it additionally requires (a) localizing the target among a potentially large number of distractors and (b) segmenting the detected object.
While a few groups have started working on variants of this task \cite{Caelles2017, Shaban2017a}, no commonly employed benchmark has emerged yet.

Our contributions are as follows:\vspace{-8pt}
\begin{itemize}
    \item We propose a new benchmark dataset: ``cluttered Omniglot'' (Fig.~\ref{fig:teaser}A). It is based on simple components -- characters from Omniglot \cite{Lake2015} -- yet turns out to be hard for current state-of-the-art computer vision components. We publish the dataset, the code and our models.\footnote{\scriptsize\texttt{https://github.com/michaelisc/cluttered-omniglot}}
    \item We present a baseline for one-shot segmentation on cluttered Omniglot. It combines two principled yet simple components: a Siamese network for object detection and a U-net for segmentation (Fig.~\ref{fig:teaser}B).
    \item We identify clutter as a substantial problem for current computer vision systems and investigate it using various oracles -- models with access to some ground truth information. Although the statistical complexity of the objects in cluttered Omniglot is low -- color alone completely identifies each instance --, the dead leaves environment creates difficulties for both detection and segmentation due to the similar foreground and background statistics.
    \item We propose to solve this task by a form of object-based attention: we first generate and segment multiple object proposals, then mask out background and finally decide among the ``cleaned-up'' objects (Fig.~\ref{fig:teaser}C). We show that this approach, which we call \textit{MaskNet}, improves both segmentation and localization.
\end{itemize}

Our paper is structured as follows: We start by describing the cluttered Omniglot dataset (Sec.~\ref{sec:dataset}), then explain our Siamese U-net baseline (Sec.~\ref{sec:baseline}) and MaskNet, our improved architecture (Sec.~\ref{sec:refined}), as well as the oracles we use (Sec.~\ref{sec:oracles}). We then present our experimental results (Sec.~\ref{sec:results}), discuss related work (Sec.~\ref{sec:related_work}) and conclude (Sec.~\ref{sec:conclusions}).

\section{Cluttered Omniglot}
\label{sec:dataset}

\begin{figure*}[ht]
\vskip 0.1in
\begin{center}
\centerline{\includegraphics[width=\textwidth]{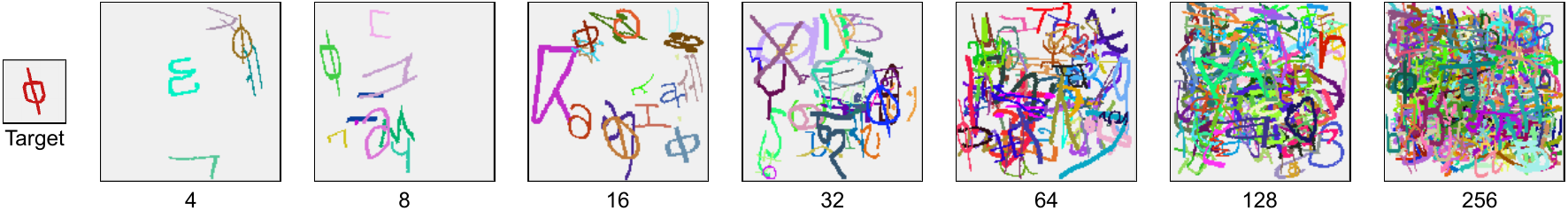}}
\caption{Multiple \textit{scenes} form \textit{cluttered Omniglot} with a common \textit{target} and varying amounts of clutter defined by the numbers of characters in each scene.}
\label{fig:dataset}
\end{center}
\vskip -0.2in
\end{figure*}

{\it Cluttered Omniglot} is a visual search task: the goal is to find a previously unseen target character in a cluttered scene and to produce a pixelwise segmentation (Fig.~\ref{fig:teaser}A).
It is based on the Omniglot dataset \cite{Lake2015}, which we chose for two reasons:
First, it is a popular and well-studied dataset for one-shot learning.
Second, the statistics of the individual objects in Omniglot are relatively simple. Nevertheless, we show below that cluttered Omniglot presents a serious challenge to convolutional neural networks.
Thus, we think of this dataset as the essence of the clutter problem.

Each sample in the dataset consists of three images: a target, a scene and a segmentation map. {\it Targets} are individual characters from Omniglot, rescaled to $32 \times 32$ pixels and colored in a random RGB color. {\it Scenes} are $96 \times 96$ pixel collages of multiple (4--256) randomly drawn Omniglot characters, one of which is the target (Fig.~\ref{fig:dataset}). The characters are sequentially ``dropped'' into the image like dead leaves, occluding any characters previously drawn at the same pixel locations. Each character is placed at a random location, has a random RGB color and is transformed with a random affine transformation of up to $20^{\circ}$ rotation, $10^{\circ}$ shearing and scaling between 16 and 64~pixels. At the end, a random instance of the target character is added. This instance is always fully visible and not occluded. We specifically avoid occlusion of the target instance, so we do not confound the effect of visual clutter with that of occlusion.

We split the dataset into three splits: training, validation and one-shot. As in the original work on Omniglot \cite{Lake2015}, we use the {\it background} set for training and validation, while we use the {\it evaluation} set for testing one-shot performance. For simplicity, we use only the first ten drawers in each alphabet for the training set and the other ten drawers for the validation and one-shot sets.

The difficulty of this task depends on the number of distractors \cite{Wolfe1998}. We show below (Section~\ref{sec:results/baseline}) that our baseline scores a close-to-perfect Intersection over Union (IoU) for the easiest version with just four distractors, similar to the accuracies of high-performing architectures designed for one-shot discrimination on Omniglot \cite{Koch2015a, Vinyals2016,Snell2017, Triantafillou2017a, Shyam2017}. In contrast, performance drops below 40\% IoU for the hardest version with 256 distractors.

For each difficulty level, we generate a training set consisting of 2 million samples and validation and one-shot sets consisting of 10,000 samples each. Note that the entire dataset is generated using a total of 9640 (6590) character instances for the training (one-shot) set.

\section{Baseline: Siamese U-net}
\label{sec:baseline}

Intuitively, the one-shot segmentation task can be broken down into two steps: detect the target in the scene and segment it. 
We implement a baseline that performs the detection part with a Siamese net applied in sliding windows over the scene to produce a heat map of candidate locations (Fig.~\ref{fig:architectures}A).
The segmentation mask is then generated by a deconvolutional net with skip connections from the encoder.

\subsection{Encoder}
\label{sec:baseline/encoder}

The encoder is inspired by Siamese networks.
It consists of two parallel fully convolutional neural networks that process the target ($32 \times 32 \times 3$) and the scene image ($96 \times 96 \times 3$), respectively (Fig.~\ref{fig:architectures}A).
All convolutions use $3 \times 3$ kernels with ``same'' padding, followed by layer normalization \cite{Ba2016b} and ReLUs.
An exception is made in the last two layers, which use $2 \times 2$ and $1 \times 1$ kernels respectively (the size of the feature maps of the target encoder in these layers) (Fig.~\ref{fig:architectures}C).
Before each but the first convolution, the image is downsampled by a factor of two using average pooling.
This architecture produces an embedding of the target in form of a 384-dimensional vector ($1 \times 1$ spatially).
The scene image is processed analogously.
To retain a higher resolution in the last layer, we do not use downsampling in the last two layers of the scene encoder. 
Instead we us a dilation factor of 2 for the convolutions in the second-to-last layer.
This results in a $12 \times 12$ pixel encoding with -- as for the target -- 384 feature maps.

Although the encoder is inspired by Siamese networks, we found in initial experiments that untying the weights improves performance and therefore do not use weight sharing between the two paths \citep[see also][]{Bertinetto2016}.
This result could potentially be attributed to the differing statistics of the clean target and the cluttered scene image.

\subsection{Target matching}

To get an estimate of the target's location in the scene, we compute the cosine similarity in the embedding space given by the encoder. We do so by taking the pixelwise inner product of the scene embedding with that of the target (Fig.~\ref{fig:architectures}C), which is implemented by a $1 \times 1$ convolution using the target embedding as the filter.
This step can be thought of as applying a Siamese network in sliding windows over the scene image (with a stride of 8, the stride of the final layer of the scene encoder). The output is a $12 \times 12$ heatmap, which can be seen as a (subsampled) pixel-level likelihood that the target is at a given location within the scene. 

This heatmap does not contain any information about what the target is. To inform the decoder about the target that should be segmented, we compute the outer tensor product of the heatmap with the target embedding. Thus, the final output of the matching step is a $12 \times 12 \times 384$ tensor, which encodes at each location the direction of the target in embedding space, weighted by how likely the encoder considers the target to be at that location. As all other layers, this output is normalized using layer normalization.

\subsection{Decoder}

The segmentation part of our baseline model is inspired by the U-net architecture \cite{Ronneberger2015a}. The decoder is essentially a mirror image of the encoder: six convolutional layers with $3 \times 3$ kernels and ``same'' padding, followed by layer normalization, ReLU and -- for the third, fourth and fifth layer -- nearest neighbor upsampling by a factor of two to incrementally increase the image size to the original $96 \times 96$ pixels (Fig.~\ref{fig:architectures}C). The input to each convolutional layer in the decoder is the concatenation of the previous layer's output and the output of the corresponding layer in the encoder (skip connections). The final layer of the decoder outputs two feature maps, which are combined into a segmentation map by taking the pixelwise softmax.

\subsection{Training}
\label{sec:baseline/training}

During training, we minimize the binary cross-entropy between the ground truth segmentation and the network's prediction.
The cross-entropy is computed pixelwise and averaged across all pixels.
The weights are initialized randomly from a Gaussian distribution following the MSRA initialization scheme \cite{He2015b}.
We regularize the weights using $L_2$ weight decay with a factor of~$10^{-9}$.
We train the network for 20 epochs using Adam \cite{Kingma2014} with a batch size of 250 and an initial learning rate of $5 \times 10^{-4}$.
After 10, 15 and 17 epochs, we divide the learning rate by 2.

\subsection{Evaluation}
\label{sec:baseline/evaluation}

We evaluated the baseline model using intersection over union (IoU).
Therefore the generated segmentation maps are binarized using a threshold or 0.3, which was determined to be optimal across models and datasets.

\begin{figure*}[ht!]
\vskip 0.05in
\begin{center}
\centerline{\includegraphics[width=0.95\textwidth]{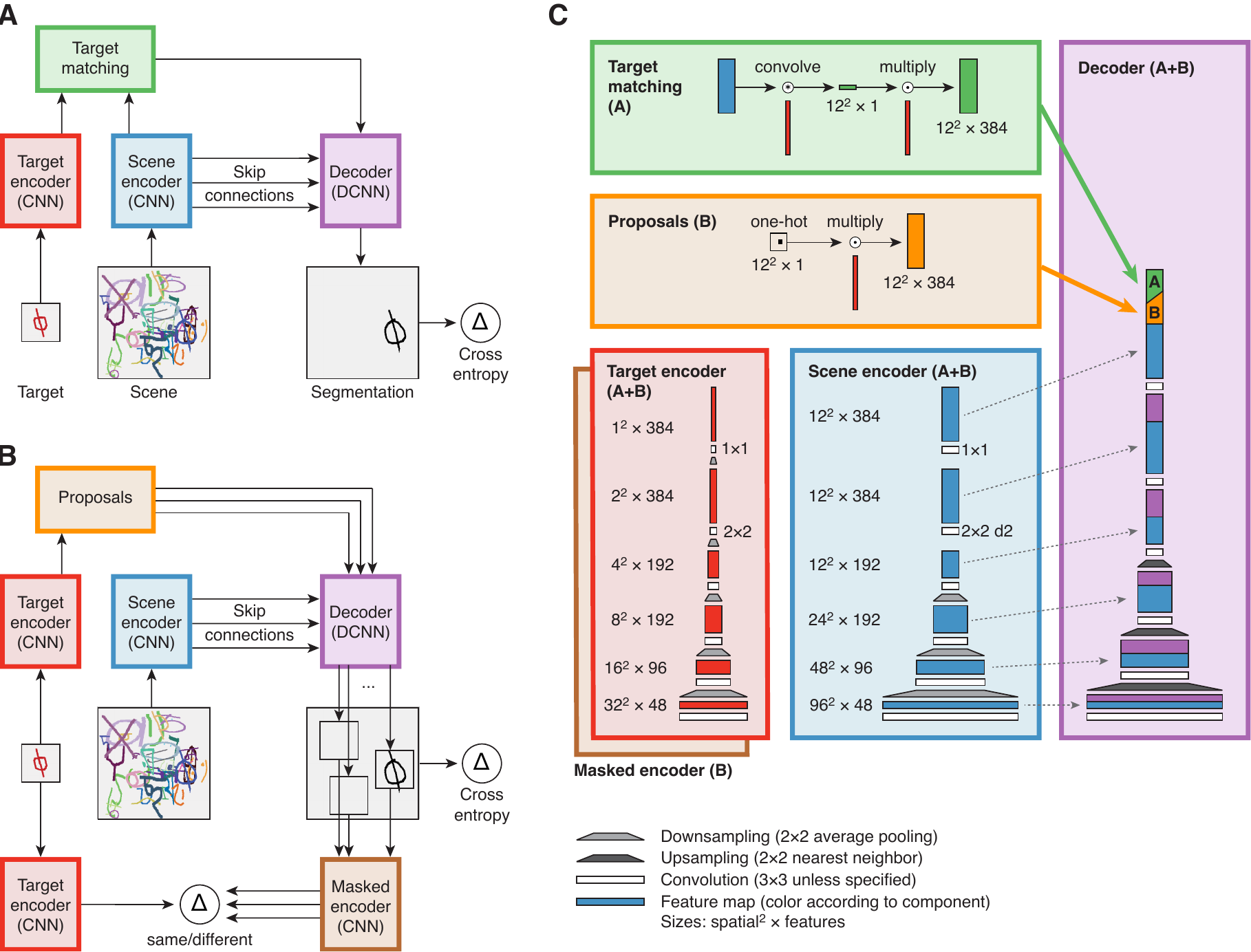}}
\caption{Architectures and details.
\textbf{A,} Siamese U-net baseline (Section~\ref{sec:baseline}).
\textbf{B,} MaskNet (Section~\ref{sec:refined}).
\textbf{C,} Close-up of the individual components, showing architecture details.
}
\label{fig:architectures}
\end{center}
\vskip -0.3in
\end{figure*}

\section{MaskNet: Segment first, decide later}
\label{sec:refined}

MaskNet (Fig.~\ref{fig:architectures}B) adds two additional processing stages to the baseline.
Instead of generating the segmentation in a single pass through the U-net, we let the decoder attend to different locations.
We branch off at the target matching stage and generate multiple object proposals with associated instance segmentations.
We then decide which of these proposals is the best match.
This last stage reduces to the one-shot multi-way discrimination task for image classification, and we solve it using a Siamese net.

\subsection{Proposal network}
\label{sec:refined/proposal}

We modify our Siamese U-net to turn it into a targeted proposal network (Fig.\ref{fig:architectures}B+C).
Its output is a set of segmentation proposals ($96 \times $96 pixels).
To this end, we modify the target matching step: instead of computing the heatmap by an inner product of target and scene embeddings, we simply set it to a one-hot map encoding a single location (Fig.\ref{fig:architectures}C, orange block).
We then use the simplest possible strategy for selecting candidate locations: sweeping all possible locations, thus generating 144 proposals (Fig.\ref{fig:architectures}B).
While there are certainly more elaborate ways of generating proposals, we opt for simplicity over efficiency.
Similar to the target matching step in the baseline network, these one-hot heatmaps are multiplied with the target embedding and normalized using layer normalization.
Thus, for each proposal, the decoder is seeded by an embedding of the target confined to a single pixel within the $12 \times 12$ spatial grid and generates a segmentation mask for the target at this location (or background if the target is not present).

\subsection{Decision stage}
\label{sec:refined/decision}

The decision stage takes multiple object proposals as input and uses a Siamese network to pick the one that most closely resembles the target (Fig.~\ref{fig:architectures}B).
This step is essentially a 144-way one-shot discrimination task.
The key ingredient here is the input: instead of just taking crops from the scene, we use the generated segmentations to mask out background clutter and perform the discrimination on ``clean'' objects (Fig.~\ref{fig:architectures}B \& Fig.~\ref{fig:teaser}C).
To do so, we binarize the segmentation proposals using a threshold of 0.3 and extend them to RGB colors by simply coloring them white.
For each proposal, we compute the center of mass of the segmentation mask and extract a $32 \times 32$ pixel crop centered on this point. We found this solution using the mask directly to perform slightly better then applying it to the image.
These crops are then fed into an encoder with the same architecture as the one used for the target (i.\,e. outputs a 384-dimensional embedding).
As in Siamese networks \cite{Koch2015a}, we use the sigmoid of a weighted sum of the L1 distance between two embeddings as a similarity measure. 
The full segmentation map corresponding to the crop that is most similar to the target is the final output.

\subsection{Training}

We train proposal network and discriminator separately, by initializing the weights (where possible) from the Siamese U-net baseline and then fine-tuning (Sec.~\ref{sec:baseline/training}).
All other weights are initialized randomly as for the baseline.
We use the same optimizer and regularization as before.
We train for five epochs, dividing the learning rate by two after two, three and four epochs, respectively.

To train the proposal network, we generate eight proposals for each training sample: four positive ones as above and four negative ones, which are drawn from random locations.
We then fine-tune encoder and decoder using the same pixelwise cross-entropy loss as above using the ground truth segmentation for the positive samples and ``background'' as the label for the negative ones.
The initial learning rate is set to $5 \times 10^{-5}$ and the batch size is 50.

To train the discriminator, we fix the target encoder, train the encoder for the segmented patches by initializing with the weights of the target encoder and fine-tuning, and train the weights for the weighted $L_1$ distance.
For each training sample, we generate four segmentation proposals: one centered at one of the four locations around the center of mass of the target and three at other random positions.
We minimize the binary cross-entropy of the same/different task for each proposal.
The initial learning rate is set to $2.5 \times 10^{-4}$ and the batch size is 250.

\subsection{Evaluation}
\label{sec:refined/evaluation}

To evaluate MaskNet, we use intersection over union (IoU) as for the baseline.
As before, we apply a threshold of 0.3 to the predicted segmentation mask.
In addition, we evaluate the localization accuracy of the network independent of the quality of the generated segmentation masks.
To do so, we use the center of mass of the chosen segmentation proposal as the prediction of the target's location.
We count all predictions that are within five pixels of the ground truth location (also center of mass) as correct and report localization accuracy in percent correct.

\section{Oracles}
\label{sec:oracles}

We evaluate two oracles that have access to ground truth segmentation masks of all characters in the scene.
Being able to define such oracles is a useful feature of cluttered Omniglot, which allows us to test the quality of individual model components.

\subsection{Pre-segmented discriminator}

The {\it pre-segmented discriminator} operates on individual characters that have been pre-segmented and cropped to the same size as the target.
Specifically, we use the fact that the characters are uniformly colored to segment each character and extract a $32 \times 32$ pixel crop centered on its center of mass.
The task of this oracle is the same as for the decision step of MaskNet (Sec.~\ref{sec:refined/decision}) and can be reduced to the widely used one-shot multi-way discrimination, hence the name {\it discriminator}.
We implement it by a Siamese network using the same encoder as before (Sec.~\ref{sec:baseline/encoder}) comparing the generated embeddings with a weighted $L_1$ distance, followed by a sigmoid \cite{Koch2015a}.
The pre-segmented discriminator lets us assess the additional difficulty (if any) introduced by (a) the random affine transformations in cluttered Omniglot and (b) the potentially large number of candidate characters to decide among.

\subsection{Cluttered discriminator}

The {\it cluttered discriminator} does not pre-segment characters.
Instead it takes the same crops as the pre-segmented discriminator, but keeps the cluttered background intact.
The rest is identical to the pre-segmented discriminator.
Thus, the cluttered discriminator performs the one-shot multi-way discrimination on cluttered crops.
By comparing its performance to that of the pre-segmented version, we can directly assess the effect of clutter on discrimination.

\begin{figure*}[t!]
\vskip 0.1in
\begin{center}
\centerline{\includegraphics[width=\textwidth]{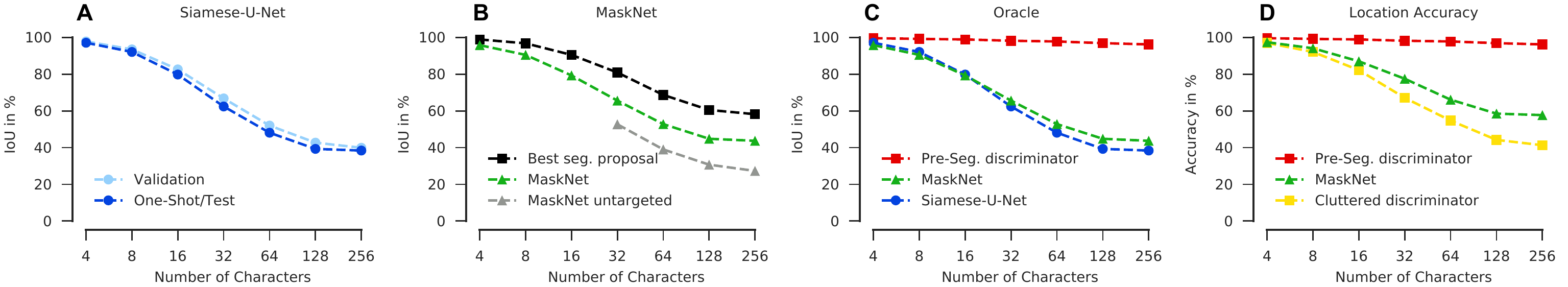}}
\caption{Performance of various model architectures and oracles on cluttered Omniglot. Performance is measured as intersection over union (IoU) for segmentation (A--C) or localization accuracy (D); higher is better. All results (except A) are measured on the one-shot sets.
\textbf{A,} IoU of the Siamese-U-Net on validation (light blue) and one-shot set (dark blue).
\textbf{B,} MaskNet with targeted (green) and un-targeted proposals (grey) and the best segmentations generated by the proposal network (black).
\textbf{C,} Comparison of Siamese-U-Net (blue), MaskNet (green) and an oracle: the pre-segmented discriminator (red), which has access to ground truth locations and segmentation masks of all characters (but not to class labels).
\textbf{D,} Localization accuracy of MaskNet (green) in comparison to the cluttered (yellow) and the pre-segmented discriminator (red).}
\label{fig:performance_comparison}
\end{center}
\vskip -0.2in
\end{figure*}

\subsection{Training}

We train both discriminators by minimizing the binary cross-entropy in the same/different task.
In each training step, four crops are sampled: one containing the target and three randomly selected ones.
Each crop is compared with the target and the average cross-entropy is computed.
Initialization, regularization and optimization are done in the same way as for the baseline (Sec.~\ref{sec:baseline/training}).
A batch size of 250 and an initial learning rate of $5 \times 10^{-4}$ are chosen.
Like the baseline, the discriminators are trained for 20 epochs and the learning rate is divided by 2 after epochs 10, 15 and 17.

\subsection{Evaluation}

We evaluate the pre-segmented discriminator using the same two metrics used for MaskNet: IoU and localization accuracy.
To evaluate IoU, we use the ground truth segmentations associated with the best-matching crop.
Due to the access to ground truth segmentations, IoU is equivalent to the percentage of correct decisions in the discrimination task.
To evaluate localization accuracy, we take the same measure as for MaskNet: The Euclidean distance between the center of each crop and the true location of the target thresholded at 5~pixels.
For the cluttered discriminator, we evaluate only localization accuracy.

\section{Results}
\label{sec:results}

We used the same encoder and decoder architectures for all experiments.
Both consist of six convolutional layers interleaved with pooling, dilation or upsampling operations (see Fig.~\ref{fig:architectures}C and Sec.~\ref{sec:baseline/encoder}).
All comparisons between architectures are therefore independent of the expressiveness of encoder and decoder, but rely only on the different approaches to segmentation and detection.
All reported results are evaluated on the one-shot set unless specified otherwise.

\subsection{Baseline}
\label{sec:results/baseline}

We start by characterizing the difficulty of the one-shot segmentation task on cluttered Omniglot by evaluating the performance of our baseline model (Section~\ref{sec:baseline}) on both, the one-shot and the validation set across all difficulty levels.

We first consider the results on the validation set (Fig.~\ref{fig:performance_comparison}A, light blue). 
The validation set contains characters seen during training, but drawn by a different set of drawers (see Section~\ref{sec:dataset}).
For a small number of distractors, the network performs well -- as expected, because the characters are mostly isolated within the scene. Performance is above 90\% IoU, similar to discrimination performance in one-shot five-way discrimination on regular Omniglot \cite{Koch2015a, Vinyals2016, Snell2017, Triantafillou2017a, Shyam2017}.
However, performance drops substantially with increasing number of distractors ($<40\%$ for 256 distractors).

On the one-shot set -- that is, characters from alphabets not seen during training --  performance is on average only 3\% worse than validation performance (Fig.~\ref{fig:performance_comparison}A, blue), showing that the network has indeed learned the right metric to identify previously unseen letters and segment them.

\begin{table}[t]
\vspace{-6pt}
\caption{One-shot segmentation accuracy (IoU in \%) across different amounts of clutter (number of characters per image).}
\label{table:results}
\vspace{-0.25cm}
\begin{center}
\begin{tiny}
\begin{sc}
\begin{tabular}{lccccccc}
\toprule
Model & 4 & 8 & 16 & 32 & 64 & 128 & 256 \\
\midrule
Pattern Matching & 62.2 & 50.4 & 41.7 & 36.9 & 32.6 & 29.0 & 28.6\\
P-seg. discriminator & 99.6 & 99.2 & 98.9 & 98.2 & 97.8 & 96.9 & 96.2\\
Best seg. proposal & 98.9 & 96.8 & 90.5 & 80.9 & 68.7 & 60.5 & 58.2\\
\midrule
Siamese U-Net & \bf{97.1} & \bf{92.1} & \bf{79.8} & 62.4 & 48.1 & 39.3 & 38.4\\
{\bf MaskNet} & 95.8 & 90.5 & 79.3 & \bf{65.6} & \bf{52.8} & \bf{44.8} & \bf{43.7}\\
MaskN. untargeted & - & - & - & 52.7 & 39.0 & 30.7 & 27.3\\
\bottomrule
\end{tabular}
\end{sc}
\end{tiny}
\end{center}
\vskip -0.1in
\end{table}

\subsection{Clutter reduces performance more than the number of comparisons}
\label{sec:results/oracles}

The performance drop of our baseline model with increasing number of distractors could have two reasons.
First, the scenes are highly cluttered, which may cause problems for the detection of the target.
Second, the large number of comparisons may simply increase the probability of making a mistake by chance ($n$-way discrimination with large $n$).
To understand the influence of these factors, we constructed two oracles, which both have access to the ground truth locations of all characters in the scene (Sec.~\ref{sec:oracles}).
Both models extract crops centered at the location of each character in the scene and perform a discrimination task between these crops and the target.

The pre-segmented discriminator has access not only to the ground truth location but also the segmentation mask of each character, allowing it to pre-segment all crops.
The resulting task is essentially the classical one-shot $n$-way discrimination task.
The only difference is that it is a bit easier since many characters in the background are highly occluded, whereas the target is always unoccluded.
Remarkably, the performance of the pre-segmented discriminator remains above 95\% IoU even for the most cluttered scenes with 256 characters (Fig.~\ref{fig:performance_comparison}C+D, red), demonstrating that our encoder can solve the task in an uncluttered environment.

\begin{table}[t]
\vspace{-6pt}
\caption{One-shot localization accuracy (in \%) across different amounts of clutter (number of characters per image).}
\label{table:results_distance}
\vspace{-0.25cm}
\begin{center}
\begin{tiny}
\begin{sc}
\begin{tabular}{lccccccc}
\toprule
Model & 4 & 8 & 16 & 32 & 64 & 128 & 256 \\
\midrule
P-Seg. discriminator & 99.6 & 99.2 & 98.9 & 98.2 & 97.8 & 96.9 & 96.2\\
\midrule
Clutt. discriminator & 97.0 & 92.1 & 82.2 & 67.1 & 54.7 & 44.2 & 41.3\\
MaskNet & \bf{97.4} & \bf{94.1} & \bf{87.0} & \bf{77.5} & \bf{66.1} & \bf{58.5} & \bf{57.7}\\
\bottomrule
\end{tabular}
\end{sc}
\end{tiny}
\end{center}
\vskip -0.1in
\end{table}

The cluttered discriminator has access to only the ground truth locations.
It cannot segment the characters and has to perform the $n$-way discrimination on cluttered crops.
In contrast to the pre-segmented discriminatior its performance takes a substantial hit with increased clutter (Fig.~\ref{fig:performance_comparison}D, yellow). Thus we conclude that the difficulty of cluttered Omniglot arises due to clutter rather than the potentially large number of candidate characters in the scene.

\subsection{Template matching is not sufficient}

A lot of work on one-shot learning has used Omniglot, but we are not aware of any work evaluating simple approaches like template matching. As a sanity check, we implemented a template matching procedure for our task based on the pre-segmented discriminator.\footnote{We generated 9317 transformed versions of the target (11 rotations, 7 shearing angles, 11x11 x/y scales), convolved them with each segmented, binarized character and picked the best match.} Accuracy ranged from 62\% for 4 characters to 29\% for 256 characters (Table~\ref{table:results}).\footnote{For comparison: on the standard 5-way one-shot task on Omniglot, we achieved 84\% accuracy using template matching.} Despite the highly simplified setting with oracle information available, template matching performs not only worse than the pre-segmented discriminator (99$-$96\%), but even worse than our baseline on the full task (97$-$38\%). Thus, template matching is not a viable solution for (cluttered) Omniglot.

\subsection{Background masking improves performance}
\label{sec:results/full_model}

Motivated by the superb discrimination performance on pre-segmented objects, we developed MaskNet, a novel model that operates in three steps (Sec.~\ref{sec:refined}).
First, we generate a number of object proposals.
Next, we generate corresponding object segmentations which mask out the background.
In the last step, we perform discrimination on these segmented objects to decide which one to pick.
This model outperforms the baseline (Fig.~\ref{fig:performance_comparison}B+C, green line), suggesting that segmenting objects (and masking out background) before classifying them is beneficial when processing highly cluttered scenes.
Nevertheless, there is still a large margin to the performance of the pre-segmented oracle.
We investigate the reasons for this margin below.

\subsection{Quality of segmentation limits performance}
\label{sec:results/upper_bound}

A crucial feature of MaskNet (and perhaps its main weakness) is that the final discriminator can only be as good as the segmentations it receives as input.
We therefore evaluate the quality of these segmentations.
To this end, we evaluate the maximal IoU among all proposals, which is equivalent to assuming a perfect discriminator that always picks the correct character.
We find that indeed  the instance segmentations of the proposals appear to be a limiting factor: for the most cluttered scenes the proposal with the highest IoU achieves only around 60\% on average (Fig.~\ref{fig:performance_comparison}B, black).

\subsection{Targeted segmentations improve performance}
\label{sec:results/ablation}

Next, we test whether it is necessary to seed the decoder with an embedding of the target, instead of just seeding it with a location and segment the most salient character at that location.
To this end, we remove the target multiplication step from MaskNet's proposal network and simply seed the decoder with the spatial one-hot encoding (Section~\ref{sec:refined/proposal}).
Using this non-targeted proposal network instead of the targeted one reduces performance (Fig.~\ref{fig:performance_comparison}B, grey), showing that it is important to supply the decoder with information what to segment.

\subsection{Performing segmentation improves localization}
\label{sec:results/distances}

So far, we have focused our evaluation of MaskNet's performance on segmentation.
Interestingly, though, segmenting objects also helps if we are interested only in localizing the target rather than segmenting it.
To provide evidence for this claim, we compare the localization performance of MaskNet to that of the cluttered discriminator.
For the cluttered discriminator, we simply use the location of the crop it chooses as the prediction for the target's location.
For MaskNet, we use the center of mass of its predicted segmentation mask.
We then compute the localization accuracy (Sec.~\ref{sec:refined/evaluation}) of these predictions to the ground truth center of mass of the target.
Indeed, MaskNet predicts the location of the target more accurately than the cluttered discriminator (Fig.~\ref{fig:performance_comparison}D and Table.~\ref{table:results_distance}), showing that segmenting objects to mask out background clutter improves localization.

\section{Related Work}
\label{sec:related_work}

\subsection{One-shot discrimination}

One-shot learning has been explored mostly in the context of multi-way discrimination for image classification.
\citeauthor{Lake2015} \citeyearpar{Lake2015} developed the Omniglot dataset for this purpose and approach it using a generative model of stroke patterns. 
Most competing approaches learn an embedding to compute a similarity metric \cite{Koch2015a, Vinyals2016,Snell2017,Triantafillou2017a}.
\citeauthor{Bertinetto2016} \citeyearpar{Bertinetto2016} train a meta network that predicts the weights of a discriminator in a single feedforward step.
Another approach compares image parts in an iterative fashion \cite{Shyam2017}.

\subsection{Semantic/instance segmentation}

Most recent approaches to segmentation use an encoder/decoder architecture \cite{Noh2015a,Badrinarayanan2017}.
The encoders are usually high-performing architectures for image classification [e.\,g. AlexNet \cite{Krizhevsky2012}, VGG \cite{Simonyan2015}, ResNet \cite{He2016a}].
The main differences lie in the decoder design.
Where early works converted high-level representations into pixelwise labels using upsampling in combination with linear transformation \cite{Long2015} or conditional random fields \cite{Chen2014, Chen2018}, recent approaches rely on more complex decoders [DeconvNet \cite{Noh2015a}, SegNet \cite{Badrinarayanan2017}, RefineNet \cite{Lin2017}] and introduce skip connections from the encoder.
The U-net architecture \citep{Ronneberger2015a}, which uses skip connections is a particularly simple and elegant general-purpose architecture for dense labeling and image-to-image problems \citep[e.\,g.][]{Isola2016}.

More recent work focuses on multi-scale pooling \cite{Zhao2017a} and dilated convolutions \cite{Chen2017c}.
These architectures improve performance, but simplify the decoders, relying more on upsampling. While this approach works well on datasets such as MS-COCO, it renders them infeasible for segmenting on Omniglot, where characters have fine detail at the pixel level.

Our proposal network is inspired by Mask R-CNN \cite{He2017a}, which achieved state-of-the-art performance on MS-COCO by splitting object detection and instance segmentation into two consecutive steps.
Similarly, our class-agnostic segmentation is inspired by the work of \citeauthor{Hong2015a} \citeyearpar{Hong2015a} and Mask R-CNN \cite{He2017a}.
Also related is work on class-agnostic segmentation using extreme point annotations \cite{Maninis2017,Papadopoulos2017}: while these works inform the segmentation by clicks in the image, our architecture seeds the decoder with a location information at the embedding layer.

\subsection{One-shot segmentation}

One-shot segmentation has emerged only recently.
\citeauthor{Caelles2017} (\citeyear{Caelles2017}) tackle the problem of segmenting an unseen object in a video based on a single (or a few) initial labeled frame(s).
The work by \citeauthor{Shaban2017a} \citeyearpar{Shaban2017a} is very similar to our approach, except that they use logistic regression with a large stride and upsampling for the decoder and tackle Pascal VOC \cite{Everingham2012}.

\subsection{Other related problems}

Co-segmentation \cite{Faktor2013,Quan2016,Sharma2017} is somewhat related to one-shot segmentation, as the common object in multiple images has to be segmented. However, objects are typically quite salient (otherwise the problem is not well defined). We can think of cluttered Omniglot as an asymmetric co-segmentation problem with one object-centered and one scene image.

Apparel recognition \cite{HadiKiapour2015, Zhao2016,Cheng2017b} and particular object retrieval \cite{Razavian2014a,Tolias2016,Li2017d,Simeoni2017} are related in the sense that the goal is to find objects specified by one image in other images.
However, both problems are primarily about image retrieval rather than segmentation of objects within these images.
One exception is the work of \citeauthor{Zhao2016} \citeyearpar{Zhao2016} in which co-segmentation is performed on pieces of clothing.


\section{Conclusions}
\label{sec:conclusions}

We explored one-shot segmentation in cluttered Omniglot and found increasing clutter to quickly diminish performance even though characters can be easily identified by color. Thus clutter is a serious problem for current state-of-the-art CNN architectures.
As a first step towards solving this problem, we showed that segmenting objects first improves detection when scenes are cluttered. We aimed for a proof of principle and thus used the simplest model possible, which performs only one iteration of segmentation and then decides directly based upon this first segmentation. Fully recurrent architectures that iteratively refine detection and segmentation by cycling through this process multiple times could lead to even larger performance gains.

As we focus on the role of clutter, we specifically designed cluttered Omniglot to have relatively simple object statistics but various levels of clutter. 
An interesting avenue for future work would be to specifically investigate cluttered image regions in real-world datasets such as Pascal VOC, MS-COCO or ADE20k. 
Both, the task and our MaskNet architecture should be directly applicable to these datatsets, for instance by searching for unseen object categories in natural scenes could be done by replacing our encoder by a state-of-the-art ImageNet classifier.

\section*{Acknowledgements}

This work was supported by the German Research Foundation (DFG) through Collaborative Research Center (CRC 1233) ``Robust Vision'' and DFG grant EC 479/1-1, and by the Advanced Research Projects Activity (IARPA) via Department of Interior/Interior Business Center (DoI/IBC) contract number D16PC00003. The U.S. Government is authorized to reproduce and distribute reprints for Governmental purposes notwithstanding any copyright annotation thereon. Disclaimer: The views and conclusions contained herein are those of the authors and should not be interpreted as necessarily representing the official policies or endorsements, either expressed or implied, of IARPA, DoI/IBC, or the U.S. Government.




\begin{thebibliography}{42}
\providecommand{\natexlab}[1]{#1}
\providecommand{\url}[1]{\texttt{#1}}
\expandafter\ifx\csname urlstyle\endcsname\relax
  \providecommand{\doi}[1]{doi: #1}\else
  \providecommand{\doi}{doi: \begingroup \urlstyle{rm}\Url}\fi

\bibitem[Ba et~al.(2016)Ba, Kiros, and Hinton]{Ba2016b}
Ba, J.~L., Kiros, J.~R., and Hinton, G.~E.
\newblock Layer {Normalization}.
\newblock \emph{arXiv:1607.06450 [cs, stat]}, 2016.
\newblock URL \url{http://arxiv.org/abs/1607.06450}.

\bibitem[Badrinarayanan et~al.(2017)Badrinarayanan, Kendall, and
  Cipolla]{Badrinarayanan2017}
Badrinarayanan, V., Kendall, A., and Cipolla, R.
\newblock {SegNet}: {A} {Deep} {Convolutional} {Encoder}-{Decoder}
  {Architecture} for {Image} {Segmentation}.
\newblock \emph{TPAMI}, 39\penalty0 (12):\penalty0 2481--2495, 2017.
\newblock \doi{10.1109/TPAMI.2016.2644615}.

\bibitem[Bertinetto et~al.(2016)Bertinetto, Henriques, Valmadre, Torr, and
  Vedaldi]{Bertinetto2016}
Bertinetto, L., Henriques, J.~F., Valmadre, J., Torr, P., and Vedaldi, A.
\newblock Learning feed-forward one-shot learners.
\newblock In \emph{NIPS}, pp.\  523--531. 2016.

\bibitem[Caelles et~al.(2017)Caelles, Maninis, Pont-Tuset, Leal-Taix{\'e},
  Cremers, and Van~Gool]{Caelles2017}
Caelles, S., Maninis, K.-K., Pont-Tuset, J., Leal-Taix{\'e}, L., Cremers, D.,
  and Van~Gool, L.
\newblock One-shot video object segmentation.
\newblock In \emph{{CVPR}}, 2017.

\bibitem[Chen et~al.(2014)Chen, Papandreou, Kokkinos, Murphy, and
  Yuille]{Chen2014}
Chen, L.-C., Papandreou, G., Kokkinos, I., Murphy, K., and Yuille, A.~L.
\newblock Semantic {Image} {Segmentation} with {Deep} {Convolutional} {Nets}
  and {Fully} {Connected} {CRFs}.
\newblock \emph{arXiv:1412.7062 [cs]}, 2014.
\newblock URL \url{http://arxiv.org/abs/1412.7062}.

\bibitem[Chen et~al.(2017)Chen, Papandreou, Schroff, and Adam]{Chen2017c}
Chen, L.-C., Papandreou, G., Schroff, F., and Adam, H.
\newblock Rethinking {Atrous} {Convolution} for {Semantic} {Image}
  {Segmentation}.
\newblock \emph{arXiv:1706.05587 [cs]}, 2017.
\newblock URL \url{http://arxiv.org/abs/1706.05587}.

\bibitem[Chen et~al.(2018)Chen, Papandreou, Kokkinos, Murphy, and
  Yuille]{Chen2018}
Chen, L.~C., Papandreou, G., Kokkinos, I., Murphy, K., and Yuille, A.~L.
\newblock {DeepLab}: {Semantic} {Image} {Segmentation} with {Deep}
  {Convolutional} {Nets}, {Atrous} {Convolution}, and {Fully} {Connected}
  {CRFs}.
\newblock \emph{TPAMI}, 2018.
\newblock \doi{10.1109/TPAMI.2017.2699184}.

\bibitem[Cheng et~al.(2017)Cheng, Wu, Liu, and Hua]{Cheng2017b}
Cheng, Z.-Q., Wu, X., Liu, Y., and Hua, X.-S.
\newblock Video2shop: {Exact} {Matching} {Clothes} in {Videos} to {Online}
  {Shopping} {Images}.
\newblock In \emph{CVPR}, pp.\  4048--4056, 2017.

\bibitem[Everingham et~al.(2012)Everingham, Van~Gool, Williams, Winn, and
  Zisserman]{Everingham2012}
Everingham, M., Van~Gool, L., Williams, C. K.~I., Winn, J., and Zisserman, A.
\newblock The {PASCAL} {V}isual {O}bject {C}lasses {C}hallenge 2012
  {(VOC2012)}, 2012.
\newblock URL
  \url{http://www.pascal-network.org/challenges/VOC/voc2012/workshop/index.html}.

\bibitem[Faktor \& Irani(2013)Faktor and Irani]{Faktor2013}
Faktor, A. and Irani, M.
\newblock Co-segmentation by {Composition}.
\newblock In \emph{ICCV}, pp.\  1297--1304, 2013.
\newblock URL \url{http://ieeexplore.ieee.org/document/6751271/}.

\bibitem[Hadi~Kiapour et~al.(2015)Hadi~Kiapour, Han, Lazebnik, Berg, and
  Berg]{HadiKiapour2015}
Hadi~Kiapour, M., Han, X., Lazebnik, S., Berg, A.~C., and Berg, T.~L.
\newblock Where to buy it: {Matching} street clothing photos in online shops.
\newblock In \emph{ICCV}, pp.\  3343--3351, 2015.
\newblock URL
  \url{http://www.cv-foundation.org/openaccess/content_iccv_2015/html/Kiapour_Where_to_Buy_ICCV_2015_paper.html}.

\bibitem[He et~al.(2015)He, Zhang, Ren, and Sun]{He2015b}
He, K., Zhang, X., Ren, S., and Sun, J.
\newblock Delving deep into rectifiers: {Surpassing} human-level performance on
  imagenet classification.
\newblock In \emph{ICCV}, pp.\  1026--1034, 2015.

\bibitem[He et~al.(2016)He, Zhang, Ren, and Sun]{He2016a}
He, K., Zhang, X., Ren, S., and Sun, J.
\newblock Deep residual learning for image recognition.
\newblock In \emph{CVPR}, pp.\  770--778, 2016.

\bibitem[He et~al.(2017)He, Gkioxari, Doll{\'a}r, and Girshick]{He2017a}
He, K., Gkioxari, G., Doll{\'a}r, P., and Girshick, R.
\newblock Mask {R}-{CNN}.
\newblock In \emph{ICCV}, pp.\  2980--2988, October 2017.
\newblock \doi{10.1109/ICCV.2017.322}.

\bibitem[Hong et~al.(2015)Hong, Noh, and Han]{Hong2015a}
Hong, S., Noh, H., and Han, B.
\newblock Decoupled {Deep} {Neural} {Network} for {Semi}-supervised {Semantic}
  {Segmentation}.
\newblock In \emph{NIPS}, pp.\  1495--1503. 2015.

\bibitem[Isola et~al.(2016)Isola, Zhu, Zhou, and Efros]{Isola2016}
Isola, P., Zhu, J.-Y., Zhou, T., and Efros, A.~A.
\newblock Image-to-{Image} {Translation} with {Conditional} {Adversarial}
  {Networks}.
\newblock \emph{arXiv:1611.07004 [cs]}, 2016.
\newblock URL \url{http://arxiv.org/abs/1611.07004}.

\bibitem[Kingma \& Ba(2014)Kingma and Ba]{Kingma2014}
Kingma, D.~P. and Ba, J.
\newblock Adam: {A} {Method} for {Stochastic} {Optimization}.
\newblock \emph{arXiv:1412.6980 [cs]}, 2014.
\newblock URL \url{http://arxiv.org/abs/1412.6980}.

\bibitem[Koch et~al.(2015)Koch, Zemel, and Salakhutdinov]{Koch2015a}
Koch, G., Zemel, R., and Salakhutdinov, R.
\newblock Siamese {Neural} {Networks} for {One}-shot {Image} {Recognition} -
  oneshot1.pdf.
\newblock \emph{ICML}, 2015.

\bibitem[Krizhevsky et~al.(2012)Krizhevsky, Sutskever, and
  Hinton]{Krizhevsky2012}
Krizhevsky, A., Sutskever, I., and Hinton, G.~E.
\newblock Imagenet classification with deep convolutional neural networks.
\newblock In \emph{NIPS}, pp.\  1097--1105, 2012.

\bibitem[Lake et~al.(2015)Lake, Salakhutdinov, and Tenenbaum]{Lake2015}
Lake, B.~M., Salakhutdinov, R., and Tenenbaum, J.~B.
\newblock Human-level concept learning through probabilistic program induction.
\newblock \emph{Science}, 350\penalty0 (6266):\penalty0 1332--1338, 2015.
\newblock URL \url{http://science.sciencemag.org/content/350/6266/1332}.

\bibitem[Li et~al.(2017)Li, Wang, Li, Agustsson, Berent, Gupta, Sukthankar, and
  Van~Gool]{Li2017d}
Li, W., Wang, L., Li, W., Agustsson, E., Berent, J., Gupta, A., Sukthankar, R.,
  and Van~Gool, L.
\newblock {WebVision} {Challenge}: {Visual} {Learning} and {Understanding}
  {With} {Web} {Data}.
\newblock \emph{arXiv:1705.05640 [cs]}, 2017.
\newblock URL \url{http://arxiv.org/abs/1705.05640}.

\bibitem[Lin et~al.(2017)Lin, Milan, Shen, and Reid]{Lin2017}
Lin, G., Milan, A., Shen, C., and Reid, I.
\newblock Refinenet: {Multi}-path refinement networks for high-resolution
  semantic segmentation.
\newblock In \emph{{CVPR}}, 2017.

\bibitem[Long et~al.(2015)Long, Shelhamer, and Darrell]{Long2015}
Long, J., Shelhamer, E., and Darrell, T.
\newblock Fully convolutional networks for semantic segmentation.
\newblock In \emph{{CVPR}}, pp.\  3431--3440, 2015.

\bibitem[Maninis et~al.(2017)Maninis, Caelles, Pont-Tuset, and
  Van~Gool]{Maninis2017}
Maninis, K.-K., Caelles, S., Pont-Tuset, J., and Van~Gool, L.
\newblock Deep {Extreme} {Cut}: {From} {Extreme} {Points} to {Object}
  {Segmentation}.
\newblock \emph{arXiv:1711.09081 [cs]}, 2017.
\newblock URL \url{http://arxiv.org/abs/1711.09081}.

\bibitem[Noh et~al.(2015)Noh, Hong, and Han]{Noh2015a}
Noh, H., Hong, S., and Han, B.
\newblock Learning deconvolution network for semantic segmentation.
\newblock In \emph{{ICCV}}, pp.\  1520--1528, 2015.

\bibitem[Papadopoulos et~al.(2017)Papadopoulos, Uijlings, Keller, and
  Ferrari]{Papadopoulos2017}
Papadopoulos, D.~P., Uijlings, J.~R., Keller, F., and Ferrari, V.
\newblock Extreme clicking for efficient object annotation.
\newblock In \emph{{ICCV}}, 2017.

\bibitem[Quan et~al.(2016)Quan, Han, Zhang, and Nie]{Quan2016}
Quan, R., Han, J., Zhang, D., and Nie, F.
\newblock Object co-segmentation via graph optimized-flexible manifold ranking.
\newblock In \emph{{CVPR}}, pp.\  687--695, 2016.

\bibitem[Razavian et~al.(2014)Razavian, Azizpour, Sullivan, and
  Carlsson]{Razavian2014a}
Razavian, A.~S., Azizpour, H., Sullivan, J., and Carlsson, S.
\newblock {CNN} features off-the-shelf: an astounding baseline for recognition.
\newblock In \emph{{CVPR} {Workshops}}, pp.\  512--519, 2014.

\bibitem[Ronneberger et~al.(2015)Ronneberger, Fischer, and
  Brox]{Ronneberger2015a}
Ronneberger, O., Fischer, P., and Brox, T.
\newblock U-{Net}: {Convolutional} {Networks} for {Biomedical} {Image}
  {Segmentation}.
\newblock In \emph{Medical {Image} {Computing} and {Computer}-{Assisted}
  {Intervention}}, pp.\  234--241. Springer, 2015.
\newblock URL
  \url{https://link.springer.com/chapter/10.1007/978-3-319-24574-4_28}.

\bibitem[Shaban et~al.(2017)Shaban, Bansal, Liu, Essa, and Boots]{Shaban2017a}
Shaban, A., Bansal, S., Liu, Z., Essa, I., and Boots, B.
\newblock One-{Shot} {Learning} for {Semantic} {Segmentation}.
\newblock \emph{{BMVC}}, 2017.

\bibitem[Sharma(2017)]{Sharma2017}
Sharma, A.
\newblock One {Shot} {Joint} {Colocalization} and {Cosegmentation}.
\newblock \emph{arXiv:1705.06000 [cs]}, 2017.
\newblock URL \url{http://arxiv.org/abs/1705.06000}.

\bibitem[Shyam et~al.(2017)Shyam, Gupta, and Dukkipati]{Shyam2017}
Shyam, P., Gupta, S., and Dukkipati, A.
\newblock Attentive {Recurrent} {Comparators}.
\newblock \emph{arXiv:1703.00767 [cs]}, 2017.
\newblock URL \url{http://arxiv.org/abs/1703.00767}.

\bibitem[Sim{\'e}oni et~al.(2017)Sim{\'e}oni, Iscen, Tolias, Avrithis, and
  Chum]{Simeoni2017}
Sim{\'e}oni, O., Iscen, A., Tolias, G., Avrithis, Y., and Chum, O.
\newblock Unsupervised deep object discovery for instance recognition.
\newblock \emph{arXiv:1709.04725 [cs]}, 2017.
\newblock URL \url{http://arxiv.org/abs/1709.04725}.

\bibitem[Simonyan \& Zisserman(2015)Simonyan and Zisserman]{Simonyan2015}
Simonyan, K. and Zisserman, A.
\newblock Very {Deep} {Convolutional} {Networks} for {Large}-{Scale} {Image}
  {Recognition}.
\newblock \emph{{ICLR}}, 2015.
\newblock URL \url{http://arxiv.org/abs/1409.1556}.

\bibitem[Snell et~al.(2017)Snell, Swersky, and Zemel]{Snell2017}
Snell, J., Swersky, K., and Zemel, R.
\newblock Prototypical {Networks} for {Few}-shot {Learning}.
\newblock In \emph{{NIPS}}, pp.\  4080--4090. 2017.

\bibitem[Tolias et~al.(2016)Tolias, Sicre, and J{\'e}gou]{Tolias2016}
Tolias, G., Sicre, R., and J{\'e}gou, H.
\newblock Particular object retrieval with integral max-pooling of {CNN}
  activations.
\newblock \emph{{ICLR}}, 2016.
\newblock URL \url{http://arxiv.org/abs/1511.05879}.

\bibitem[Triantafillou et~al.(2017)Triantafillou, Zemel, and
  Urtasun]{Triantafillou2017a}
Triantafillou, E., Zemel, R., and Urtasun, R.
\newblock Few-{Shot} {Learning} {Through} an {Information} {Retrieval} {Lens}.
\newblock In \emph{{NIPS}}, pp.\  2252--2262. 2017.

\bibitem[Vinyals et~al.(2016)Vinyals, Blundell, Lillicrap, Wierstra, and
  {others}]{Vinyals2016}
Vinyals, O., Blundell, C., Lillicrap, T., Wierstra, D., and {others}.
\newblock Matching networks for one shot learning.
\newblock In \emph{{NIPS}}, pp.\  3630--3638, 2016.

\bibitem[Wolfe(1998)]{Wolfe1998}
Wolfe, J.~M.
\newblock Visual search.
\newblock \emph{Attention}, 1:\penalty0 13--73, 1998.

\bibitem[Zhao et~al.(2016)Zhao, Wu, Peng, and Yan]{Zhao2016}
Zhao, B., Wu, X., Peng, Q., and Yan, S.
\newblock Clothing {Cosegmentation} for {Shopping} {Images} {With} {Cluttered}
  {Background}.
\newblock \emph{Transactions on Multimedia}, 18\penalty0 (6):\penalty0
  1111--1123, 2016.
\newblock URL \url{http://ieeexplore.ieee.org/document/7423747/}.

\bibitem[Zhao et~al.(2017)Zhao, Shi, Qi, Wang, and Jia]{Zhao2017a}
Zhao, H., Shi, J., Qi, X., Wang, X., and Jia, J.
\newblock Pyramid scene parsing network.
\newblock In \emph{{CVPR}}, pp.\  2881--2890, 2017.

\bibitem[Zhou et~al.(2017)Zhou, Zhao, Puig, Fidler, Barriuso, and
  Torralba]{Zhou2017b}
Zhou, B., Zhao, H., Puig, X., Fidler, S., Barriuso, A., and Torralba, A.
\newblock Scene parsing through {ADE20k} dataset.
\newblock In \emph{{CVPR}}, 2017.

\end{thebibliography}

\bibliographystyle{icml2018}

\newpage

\end{document}